\documentclass[conference]{IEEEtran}
\IEEEoverridecommandlockouts
\usepackage{cite}
\usepackage{amsmath,amssymb,amsfonts}
\usepackage{algorithmic}
\usepackage{graphicx}
\usepackage{textcomp}
\usepackage{xcolor}

\makeatletter
\def\ps@IEEEtitlepagestyle{%
  \def\@oddfoot{\mycopyrightnotice}%
  \def\@evenfoot{}%
}
\def\mycopyrightnotice{%
  {\footnotesize Copyright notice: 979-8-3503-3239-1/23/\$31.00 © 2023 IEEE\hfill}
  \gdef\mycopyrightnotice{}
}

\def\BibTeX{{\rm B\kern-.05em{\sc i\kern-.025em b}\kern-.08em
    T\kern-.1667em\lower.7ex\hbox{E}\kern-.125emX}}
\begin{document}

\title{C-RADAR: A Centralized Deep Learning System for Intrusion Detection in Software Defined Networks\\

}

\author{\IEEEauthorblockN{Osama Mustafa}
\IEEEauthorblockA{\textit{Center of Excellence in AI (CoE-AI)} \\
\textit{Bahria University Islamabad}\\
Islamabad, Pakistan \\
muhammadosama939@gmail.com}
\and
\IEEEauthorblockN{Khizer Ali}
\IEEEauthorblockA{\textit{Center of Excellence in AI (CoE-AI)} \\
\textit{Bahria University Islamabad}\\
Islamabad, Pakistan \\
mkhizer.buic@bahria.edu.pk}
\and
\IEEEauthorblockN{Talha Naqash}
\IEEEauthorblockA{\textit{Department of Computer Science} \\
\textit{Bahria University Islamabad}\\
 Islamabad, Pakistan\\
talha.naqashtn@gmail.com}

}

\maketitle

\begin{abstract}
The popularity of Software Defined Networks (SDNs) has grown in recent years, mainly because of their ability to simplify network management and improve network flexibility. However, this also makes them vulnerable to various types of cyber attacks. SDNs work on a centralized control plane which makes them more prone to network attacks. Research has demonstrated that deep learning (DL) methods can be successful in identifying intrusions in conventional networks, but their application in SDNs is still an open research area. In this research, we propose the use of DL techniques for intrusion detection in SDNs. We measure the effectiveness of our method by experimentation on a dataset of network traffic and comparing it to existing techniques. Our results show that the DL-based approach outperforms traditional methods in terms of detection accuracy and computational efficiency. The deep learning architecture that has been used in this research is a Long Short Term Memory Network and Self-Attention based architecture i.e. LSTM-Attn which achieves an F1-score of 0.9721. Furthermore, this technique can be trained to detect new attack patterns and improve the overall security of SDNs.
\end{abstract}


\section{Introduction}
This study focuses on intrusion detection in software defined networks. Software Defined Networking (SDN) \cite{10.1145/2602204.2602219} is an important research dimension because it has the potential to revolutionize the way networks are managed and operated. In SDN, the control plane, responsible for making decisions on traffic forwarding, is separated from the data plane, which physically forwards the traffic. This separation allows for more flexibility, programmability and centralized control of the network\cite{6834762}.

SDNs and the current networking scenario has a direct relation with the emerging technologies. Artificial Intelligence applications have started to become more common especially on common user devices and as AI systems are closely dependent on large amounts of data there is a need of robust and scalable networking infrastructure that can support this technology. As due to data privacy concerns the concept of federated learning has started to become more common, we need a system that is more robust and scalable. In order to support all these emerging technologies, there are some issues in conventional networks such as customization, programmability and scalability. In conventional networks in order to do some customization, it has to be done on the firmware level. Software Defined Networks cover all these issues and offer customization and scalability\cite{6496914} A python application in the form of a software is easily deployable on an SDN controller. Software-Defined Networking (SDN) is a rapidly emerging field that is changing the way networks are managed and controlled. Instead of traditional networks, which rely on proprietary hardware and closed systems, SDNs use software to control and manage the flow of data across a network. One of the key benefits of SDNs is their flexibility and programmability. Because the control plane is separated from the data plane in SDN architecture, network administrators can easily make changes to the network without having to reconfigure hardware. This allows for more efficient and faster deployment of new services and applications. Another advantage of SDNs is the ability to automate network management tasks. By using software-based controllers, network administrators can create policies and rules that automatically manage the flow of data across the network. This can lead to more efficient use of network resources and improved security. The current networking requirements, such as the increasing number of connected devices, the need for faster and more reliable networks and the need for more secure networks, all favor the need for SDNs. With the ability to easily scale and adapt to changing requirements, SDNs can provide a more efficient and cost-effective solution for modern networks. In conclusion, Software-Defined Networking (SDN) is an emerging field that offers many benefits over traditional networking approaches. Its flexibility, programmability, and automation capabilities make it an ideal solution for meeting the current and future networking requirements. As the demand for more efficient, reliable, and secure networks continues to grow, the adoption of SDN technology is likely to increase in the future.

There are some security challenges that come along with a centralized control plane\cite{dacier2017security}. A centralized control plane in Software Defined Networking (SDN) can present a security challenge because it represents a single point of failure and a potential target for attackers. With a centralized control plane, all network decisions and configurations are made by a central controller, which can be a software or a specialized hardware. This centralization can make the controller a prime target for attackers, as a successful attack on the controller can compromise the entire network. Additionally, if the controller is compromised, an attacker can potentially gain control over the entire network and launch further attacks. Another risk is if the controller is not properly secured, an attacker can potentially gain access to sensitive network information such as IP addresses, traffic patterns and configurations. Additionally, a central controller can be a bottleneck for network traffic. If the controller goes down, it can disrupt the entire network and cause a Denial of Service (DoS) attack. To mitigate these risks, it is important to properly secure the controller and the communication between the controller and the network devices.

As a result, this research proposes a deep learning-based system for identifying intrusions in Software Defined Networks (SDN). Deep learning-based intrusion detection in Software Defined Networks (SDN) can be more robust than conventional rule-based techniques for a few reasons:
\begin{itemize}
    \item Improved accuracy: Deep learning algorithms, such as neural networks, can learn to identify patterns and anomalies in network traffic that may not be easily detected by rule-based systems. This can lead to an increase in the number of true detections and a decrease in the number of false detections
    \item Handling complex and dynamic network: SDN environments are complex and dynamic, and the traffic patterns can change over time. Conventional rule-based systems may struggle to adapt to these changes, whereas deep learning algorithms can continuously learn from new data and improve their detection capabilities.
    \item Handling new and unknown attacks: Deep learning-based systems can detect new and unknown attacks by identifying patterns that deviate from normal behavior. This is because deep learning models can learn from the data to identify patterns and anomalies that are not specified by predefined rules.
    \item Scalability: Deep learning models can be trained to handle large amounts of data and traffic, which makes it scalable to handle large networks.
\end{itemize}
Further in this paper, Section~\ref{sec:lit} presents the recent advancements in intrusion detection for SDNs, Section~\ref{sec:data} presents the details of the dataset used for experimentation, Section~\ref{sec:metho} presents the proposed methodology, Section~\ref{sec:exp} presents the experimentation and results, and finally conclusion is presented in Section~\ref{sec:conc}.

\section{Related Work} \label{sec:lit}
Researchers in the field of intelligent network intrusion detection (NID) typically use techniques such as dimensionality reduction, clustering, and classification to differentiate normal network traffic from abnormal traffic, in order to identify and detect malicious attacks \cite{cieslak2006combating,zamani2013machine}. Pervez et al. put forward a technique for combining feature selection and classification for the multi-class NSL-KDD Cup99 dataset using Support Vector Machine (SVM) and evaluated the classification accuracy of the classifiers under various feature dimensions \cite{pervez2014feature}. Sheraz employed the K Farthest Neighbor (KFN) and K Nearest Neighbor (KNN) algorithms to classify the data, and in cases where the nearest and farthest neighbors had the same class label, he used the Second Nearest Neighbor (SNN) for classification \cite{shapoorifard2017intrusion}. Similarly, Bhattacharya et al. proposed a ML technique that combines Principal Component Analysis (PCA) and Firefly algorithm. The model first applies one key coding to the IDS dataset, then reduces the dimensionality of the dataset using the hybrid PCA-Firefly algorithm, and finally uses the XGBoost algorithm to classify the reduced dataset \cite{bhattacharya2020novel}.

As deep learning has revolutionized other research areas like Natural Language Processing, and Computer Vision, therefore, its also been used recently by researchers for intrusion detection. Generally, the neural networks learn the features from the labeled dataset during the training of the model.  Torres utilized Recurrent Neural Network (RNN) to detect malicious network traffic by first converting network traffic characteristics into characters and then analyzing their temporal characteristics \cite{torres2016analysis}. Wang et al. proposed a classification algorithm for malicious software traffic using a Convolutional Neural Network (CNN) \cite{wang2017malware}. Another team conducted research on deep learning, focusing on techniques such as data simplification, dimension reduction, and classification. They proposed a Fully Convolutional Network (FCN) model as a result of their research \cite{kwon2019survey}. Tama and their team proposed an Intrusion Detection System (IDS) that uses a two-stage meta-classifier for anomaly-based detection. The authors employed a combination of feature selection techniques to obtain precise feature representations and found that it enhanced detection rates when evaluating on the NSL-KDD and UNSW-NB15 intrusion datasets \cite{tama2019tse}.   
\begin{figure}[!h]
    \includegraphics[width = 0.5\textwidth]{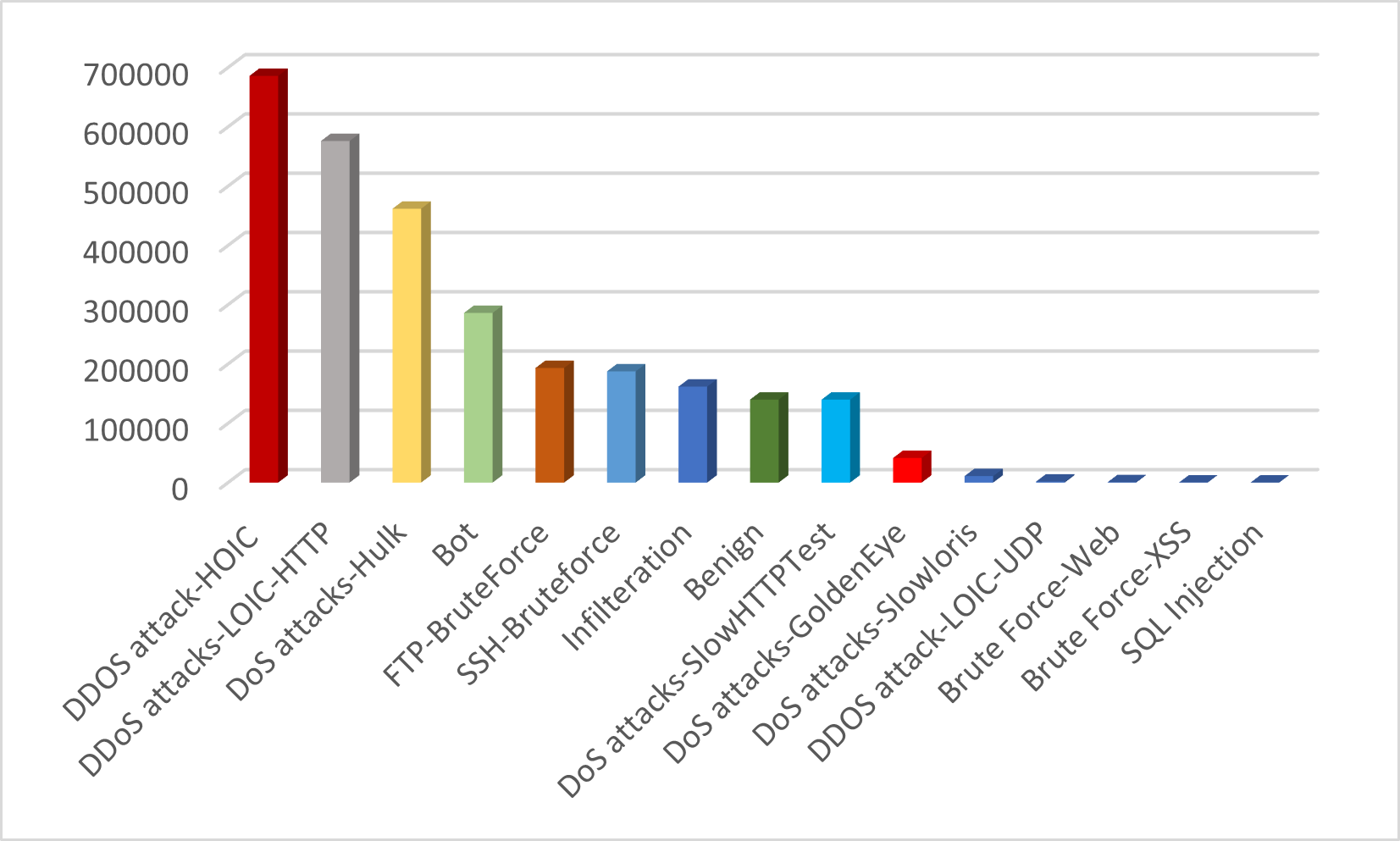}
    \caption{Histogram representing class distribution in dataset}
    \label{fig:hist}
\end{figure}
\section{Dataset} \label{sec:data}
In this study, experimentation is conducted on CSE-CIC-IDS2018\footnote{https://www.unb.ca/cic/datasets/ids-2018.html} \cite{Sharafaldin2018TowardGA}.

In this study, availability of data that is a proper representation of attack patterns is a common challenge since the organizations try to keep it internal for privacy concerns. This dataset contains the feature set for 15 different types of attack types, some of them are: Benign, SSH-Bruteforce, DoS attacks-GoldenEye, FTP-BruteForce, DoS attacks-Slowloris. The dataset used in this study consists of network traffic captures and system logs from each machine, as well as 80 characteristics extracted from the captured traffic through the use of CICFlowMeter-V3. There are total 80 features in the dataset, a few can be seen in Table I. Figure~\ref{fig:hist} illustrates the class distribution of each class from the dataset.

\begin{table}[!h]
\centering
\caption{Some feature examples from dataset}
\label{tab:my-table}
\begin{tabular}{ll}
\hline
\hline
\textbf{Feature} & \textbf{Description}                                         \\
\hline
fl\_dur          & Flow duration                                                \\
tot\_fw\_pk      & Total packets-forward direction                       \\
tot\_bw\_pk      & Total packets-backward direction                      \\
fl\_iat\_avg     & Average time for two flows                               \\
bw\_iat\_avg     & Mean time between two packets sent-backward-direction \\
\hline 
\end{tabular}
\end{table}

\subsection{Data Preprocessing}
In this study, while the availability of dataset is a challenge similarly preprocessing of available dataset is also a major challenge. This dataset contains data for multiple classes as detailed above and the data distribution is highly unbalanced. We have converted the dataset from multiclass to binary class by assigning all the attack values the label: malicious, and remaining values label: non-malicious. This problem is approached as a binary class problem considering the real world scenario where a malicious network traffic can be a combination as a hybrid attack. By training the model as a binary classifier it would be able to detect and flag any malicious network traffic flow.

This dataset is imbalanced and for balancing the dataset reduction approach has been used. There are two classes in the dataset after conversion to binary class: benign and malicious. The proportion of benign data points is greater than malicious, therefore in order to balance the dataset the number of benign data samples has been reduced in each file respectively. This prevents information loss as the model has to learn the patterns of malicious traffic. In recent work, SMOTE \cite{DBLP:journals/corr/abs-1106-1813} has been used extensively to solve the imbalance problem but comparison presented in section IV validates that utilizing SMOTE for this purpose reduces the performance on inference.

The dataset is initially available in multiple files respective to days as the attacks were conducted on different days. We concatenated all the data into a single csv file as part of pre-processing. The feature 'time-stamp' is removed from the dataset and remaining all of the features take part in training the model. In addition, the null values have been dropped from the dataset due to the small count as compared to non-null data values. The values have been scaled using a Min Max Scalar with default values. The train, validation and test distribution is: 0.7, 0.1, 0.2 respectively. 
\begin{figure*}[!h]
    \caption{System Pipeline}
    \centering
    \includegraphics[width = \textwidth]{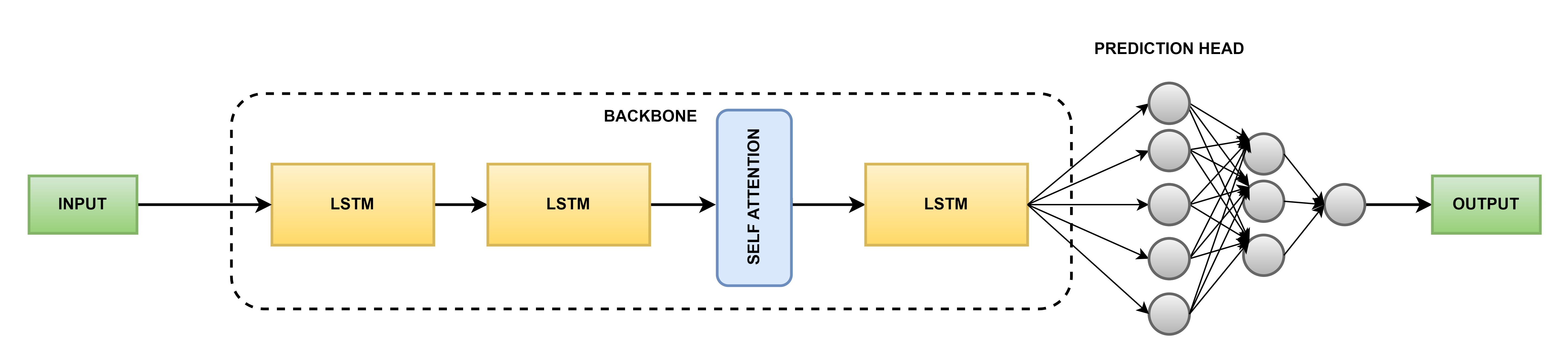}
    \label{fig:seqdiag}
\end{figure*}
\section{Methodology} \label{sec:metho}
In this study, deep learning based neural network architecture is utilized for the task of malicious traffic classification. A further class of neural networks i.e Long Short Term Memory Network (LSTM) \cite{DBLP:journals/corr/abs-1909-09586} and Self-attention \cite{DBLP:journals/corr/VaswaniSPUJGKP17} based architecture has been utilized in this study. In this research, attack patterns are complex and difficult to identify due to their slow and passive approach. LSTMs are a type of recurrent neural network \cite{DBLP:journals/corr/abs-1912-05911} that are well-suited for tasks involving sequential data. In the context of intrusion detection in software-defined networks, LSTMs can be trained to analyze network traffic and identify patterns that indicate malicious activity. They are particularly useful for this task because they are able to "remember" information from previous time steps and use it to inform their analysis of current data. Additionally, LSTMs can also handle variable-length sequences of data, which makes them well-suited for analyzing network traffic where the number of packets in a given session can vary. Using self-attention in conjunction with LSTMs can improve the performance of an intrusion detection system for several reasons.
\begin{itemize}
    \item Self-attention allows the model to focus on specific parts of the input sequence, which can be useful for identifying important features in the network traffic data that indicate malicious activity
    \item Self-attention mechanisms can learn to weight the importance of different parts of the input sequence, which can be useful in intrusion detection where some parts of the network traffic may be more important than others in determining whether an attack is occurring
    \item By using self-attention, model can understand the relationship between different parts of the input sequence, which is useful in intrusion detection as it allows the model to identify patterns of malicious activity across different parts of the network traffic data
\end{itemize}

\subsection{Architecture}
This section details the architecture of the model used in this study. We have utilized an LSTM-Attn based model in this study. As illustrated in Figure 2 the system pipeline is comprised of a backbone (for feature extraction) and a prediction head (for decision making). Table II details the summary of the architecture utilized in this study. 
\begin{table}[!h]
\centering
\caption{Model Summary}
\label{tab:my-table}
\begin{tabular}{lll}
\hline
\hline
\textbf{Layer (type)}        & \textbf{Output Shape} & \textbf{Parameters} \\
\hline
lstm (LSTM)                  & (None, 78, 256)       & 264192              \\
lstm\_1 (LSTM)               & (None, 78, 256)       & 525312              \\
Attention (SeqSelfAttention) & (None, 78, 256)       & 16449               \\
lstm\_2 (LSTM)               & (None, 128)           & 197120              \\
dense (Dense)                & (None, 512)           & 66048               \\
dense\_1 (Dense)             & (None, 256)           & 131328              \\
dense\_2 (Dense)             & (None, 1)             & 257                 \\
\hline
\end{tabular}
\end{table}
In the backbone, two Long Short Term Memory Network (LSTM) layers are stacked on top of each other, the output from this stack is forwarded as an input to the self attention layer and finally an LSTM layer on top of it. The backbone performs the task of feature extraction after which these features are passed to a prediction head for the purpose of decision making. The prediction head is a Multi Layer Perceptron (MLP) with three fully connected layers for decision making. The final layer has 1 neuron because it is a binary class classification problem. 

\subsection{Train Configuration}
This section details the training configuration followed during the training of above mentioned architecture. In LSTM block, the configuration followed can be referred to in Table III

\begin{table}[!h]
\caption{Parameter Setting in LSTM Block}
\centering
\label{tab:my-table}
\begin{tabular}{ll}
\hline
\textbf{Hyper-parameter} & \textbf{Value}  \\
\hline
units                    & 256             \\
activation               & tanh            \\
recurrent\_activation    & sigmoid         \\
use\_bias                & True            \\
kernel\_initializer      & glorot\_uniform \\
recurrent\_initializer   & orthogonal      \\
unit\_forget\_bias       & True            \\
\hline
\end{tabular}
\end{table}
Sigmoid activation is used in attention layer, relu in two dense layers and sigmoid in final output layer respectively. Equation 1-2 shows how sigmoid and relu is computed respectively. The model is trained for 75 epochs on NVIDIA Tesla P100 16GB GPU and the total training time turned out to be 9.5 hours. The rest of the train configuration is: Optimizer: "Adam", learning-rate: 0.00003, loss-function: Binary-cross-entropy and batch\_size: 512. Equation 3 shows how binary cross entropy loss is computed, where y is the label and p is predicted probability.

\begin{equation}
    \sigma(z) = \frac{1} {1 + e^{-z}}
\end{equation}

\begin{equation}
    Relu(z) = max(0, z)
\end{equation}

\begin{equation}
   -{(y\log(p) + (1 - y)\log(1 - p))}
\end{equation}

\section{Experiments and Results} \label{sec:exp}
This section details the experimentation setting, evaluation metrics used and results achieved from experimentation. 
\subsection{Evaluation Metrics}
In order to quantify the experimentation outputs and perform analysis \& comparison we have used following evaluation metrics: accuracy, F1-score, precision and recall. Equations 1-4 show how each of the evaluation metric is computed
\begin{equation}
    Accuracy = \frac{TP+TN}{TP+TN+FP+FN}
\end{equation}
\begin{equation}
    Precision = \frac{TP}{TP+FP}
\end{equation}
\begin{equation}
    Recall = \frac{TP}{TP+FN}
\end{equation}
\begin{equation}
    F1 = \frac{2*Precision*Recall}{Precision+Recall} = \frac{2*TP}{2*TP+FP+FN}
\end{equation}
We have specifically used weighted f1-score:
\begin{itemize}
    \item \textbf{F1-Score (Weighted): }The F1 score is calculated by taking the average of the F1 scores for each class, with the weight being determined by the number of instances of that class in the dataset, known as the support.
\end{itemize}

\subsection{Results and Analysis}
This section presents the results and analysis. Table IV presents the quantified experimentation outputs that were achieved as a result of rigorous experimentation conducted in this research. The proposed system outperforms SOTA with an F1-score of 0.9721 in intrusion detection for SDNs. Table V \cite{9311173} presents the performance comparison of our proposed model architecture with other SOTA models used for this task on the same dataset. The training and validation graphs in Figure 3 show that the learning improves as a function of increasing epochs. 

\begin{table*}[!h]
\centering
\caption{Performance Comparison of Different Models on CSE-CIC-IDS2018}
\label{tab:my-table}
\begin{tabular}{lllll}
\hline
\hline
\textbf{Model}               & \textbf{Accuracy} & \textbf{Precision} & \textbf{Recall} & \textbf{F1-score} \\
\hline
\textbf{Proposed: LSTM-Self-Attention} & \textbf{0.9739}   & \textbf{0.9802}    & \textbf{0.9721} & \textbf{0.9721}   \\
\cite{9311173} SVM                          & 0.9225            & 0.9261             & 0.9225          & 0.9126            \\
\cite{9311173} XGBoost                      & 0.9398            & 0.9449             & 0.9398          & 0.9340            \\
\cite{9311173} LSTM                         & 0.9375            & 0.9444             & 0.9370          & 0.9313            \\
\cite{9311173} AlexNet                      & 0.9376            & 0.9440             & 0.9369          & 0.9313            \\
\cite{9311173} miniVGGNet                   & 0.9388            & 0.9450             & 0.9384          & 0.9326            \\
\cite{9311173} SMOTE+LSTM                   & 0.9345            & 0.9431             & 0.9344          & 0.9278            \\
\cite{9311173} SMOTE+AlexNet                & 0.9324            & 0.9423             & 0.9308          & 0.9287  
           \\   
\hline         
\end{tabular}
\end{table*}

\begin{figure*}
\centering
\begin{tabular}{cc}
  \includegraphics[width=65mm]{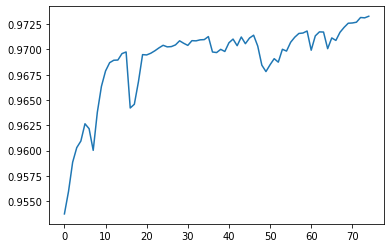} &   \includegraphics[width=65mm]{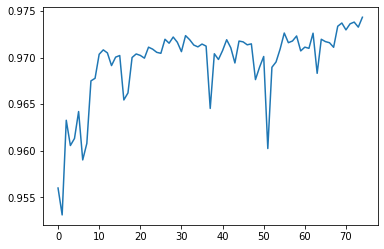} \\
(a) Accuracy & (b) Validation Accuracy \\[6pt]
 \includegraphics[width=65mm]{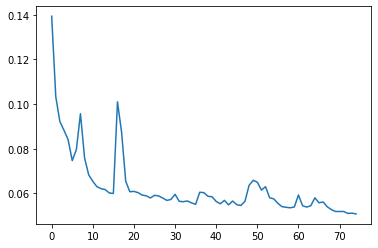} &   \includegraphics[width=65mm]{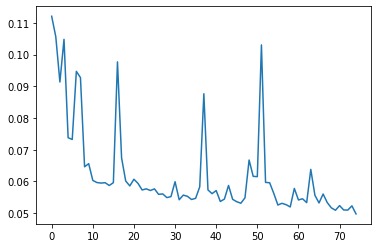} \\
(c) Loss & (d) Validation Loss \\[6pt]

\end{tabular}
\caption{Train-validation Accuracy \& Loss curves}
\label{trccurves}
\end{figure*}

Also the results show that the proposed model performs good in training as well as on validation which is a case of no overfitting. Although recent work has shown great progress in intrusion detection but results validate that our proposed methodology can be utilized for more robust and accurate intrusion detection of large scale software defined networks. The proposed methodology utilizes reduction approach instead of a synthetic data generation technique such as SMOTE for class balancing as a pre-processing step in the dataset. It is because training a model on synthetic data can prove better in training but performance drops significantly in real-world deployment. Also the results from Table V validate that the perform of LSTM and AlexNet is F1-score 0.9313 and 0.9313 respectively but after using SMOTE algorithm the performance drops with an F1-score of 0.9278 and 0.9287 respectively.

\begin{table}[!h]
\centering
\caption{Evaluation Metrics}
\label{tab:my-table}
\begin{tabular}{ll}
\hline
\hline
\textbf{Metric}     & \textbf{Value} \\
\hline
Validation-Accuracy            & 0.9739         \\
Test Set: F1-Score (Weighted) & 0.9721         \\
Test Set: Precision           & 0.9802         \\
Test Set: Recall              & 0.9721         \\     
\hline
\end{tabular}
\end{table}

\section{Conclusion} \label{sec:conc}
In conclusion, the proposed LSTM-Self-attention model outperforms state-of-the-art intrusion detection in SDN, by effectively identifying malicious traffic with an F1-score of 0.9721. This study makes a notable contribution to the area of intrusion detection systems (IDS) based on deep learning for Software Defined Networks (SDN). The model effectively leverages the temporal dependencies and attention mechanisms to accurately identify malicious network traffic. The experimentation results and analysis validate that in this case the performance drops by using synthetic data generation techniques for class balancing such as SMOTE. The proposed model can serve as an effective tool for securing software defined networks and can be further improved with additional features and optimization techniques. Overall, this research makes a significant contribution to the field of deep learning based intrusion detection systems for software defined networks.

\bibliographystyle{ieeetr}
\bibliography{References}

\end{document}